\documentclass[letterpaper, 10 pt, conference]{ieeeconf}  

\IEEEoverridecommandlockouts                              

\usepackage{graphics} 
\usepackage{epsfig} 
\usepackage{empheq}
\usepackage{times} 
\usepackage{amsmath} 
\usepackage{amssymb}  
\usepackage{bm}
\usepackage{color}
\usepackage{bbding} 
\usepackage{cite}
\usepackage{diagbox}
\usepackage[linesnumbered,ruled]{algorithm2e}
\usepackage{ulem} 
\usepackage{hyperref}
\usepackage{float}
\usepackage{booktabs}
\usepackage{multirow}
\usepackage{mathrsfs}
\usepackage[utf8]{inputenc}
\usepackage{subfigure}
\usepackage{graphicx}
\usepackage{pifont}
\usepackage{threeparttable}
\newcounter{RNum}
\renewcommand{\theRNum}{\arabic{RNum}}
\newcommand{\Remark}{\noindent\textbf{Remark}~\refstepcounter{RNum}\textbf{\theRNum}: }
\begin{document}

\title{\LARGE \bf
	ADTrack: Target-Aware Dual Filter Learning for Real-Time Anti-Dark UAV Tracking
}

\author{Bowen Li, Changhong Fu$^{*}$, Fangqiang Ding, Junjie Ye, and Fuling Lin
	\thanks{Bowen Li, Changhong Fu, Fangqiang Ding, Junjie Ye, and Fuling Lin are with the School of Mechanical Engineering, Tongji University, 201804 Shanghai, China.
		{\tt\footnotesize changhongfu@tongji.edu.cn}}%
	\thanks{$^{*}$Corresponding Author}%
}

\maketitle
\thispagestyle{empty}
\pagestyle{empty}

\begin{abstract}
Prior correlation filter (CF)-based tracking methods for unmanned aerial vehicles (UAVs) have virtually focused on tracking in the daytime. However, when the night falls, the trackers will encounter more harsh scenes, which can easily lead to tracking failure. In this regard, this work proposes a novel tracker with anti-dark function (ADTrack). The proposed method integrates an efficient and effective low-light image enhancer into a CF-based tracker. Besides, a target-aware mask is simultaneously generated by virtue of image illumination variation. The target-aware mask can be applied to jointly train a target-focused filter that assists the context filter for robust tracking. Specifically, ADTrack adopts dual regression, where the context filter and the target-focused filter restrict each other for dual filter learning. Exhaustive experiments are conducted on typical dark sceneries benchmark, consisting of 37 typical night sequences from authoritative benchmarks, \textit{i.e.}, UAVDark, and our newly constructed benchmark UAVDark70. The results have shown that ADTrack favorably outperforms other state-of-the-art trackers and achieves a real-time speed of 34 frames/s on a single CPU, greatly extending robust UAV tracking to night scenes.
\end{abstract}

\section{Introduction}

Widely applied in the field of robotics and automation, visual object tracking aims at predicting the location and size of a target object. Particularly, applying tracking methods onboard unmanned aerial vehicles (UAVs) has facilitated extensive UAV-based applications, \textit{e.g.}, collision avoidance \cite{Baca2018IROS}, autonomous aerial manipulation operations \cite{McArthur2020ICRA}, and autonomous transmission-line inspection \cite{Bian2018IROS}.

Scenarios suitable to deploy visual trackers are currently limited to daytime when the light condition is favorable and the object is easily distinguished with representative geometric and radiometric characteristics. As the night falls, the cameras fail to acquire sufficient information to complete the details of images, bringing great challenges to trackers and huge limitations to the generality and serviceability of UAV.

Compared with generic tracking scenes, visual tracking for UAV in the dark is confronted with more terrible conditions as follows: $a$) the object is apt to merge in the dark environment, making its external contour unclear; $b$) objects' color feature is usually invalid, ending up in its internal characteristics not significant; $c$) random noises appear frequently on images captured at night, severely distracting the tracker; $d$) limited computation and storage resources on UAV set barriers to real-time tracking. Due to the challenging tracking conditions above, current state-of-the-art (SOTA) methods~\cite{Li2020CVPR, Huang2019ICCV,Bertinetto2016ECCV,Dai2019CVPR,Galoogahi2017ICCV,Li2018CVPR} fail to up to scratch for UAV tracking in the dark.
\begin{figure}[!t]
\centering
\setlength{\abovecaptionskip}{-0.3cm}
\setlength{\belowcaptionskip}{-20cm}
\includegraphics[width=1\columnwidth]{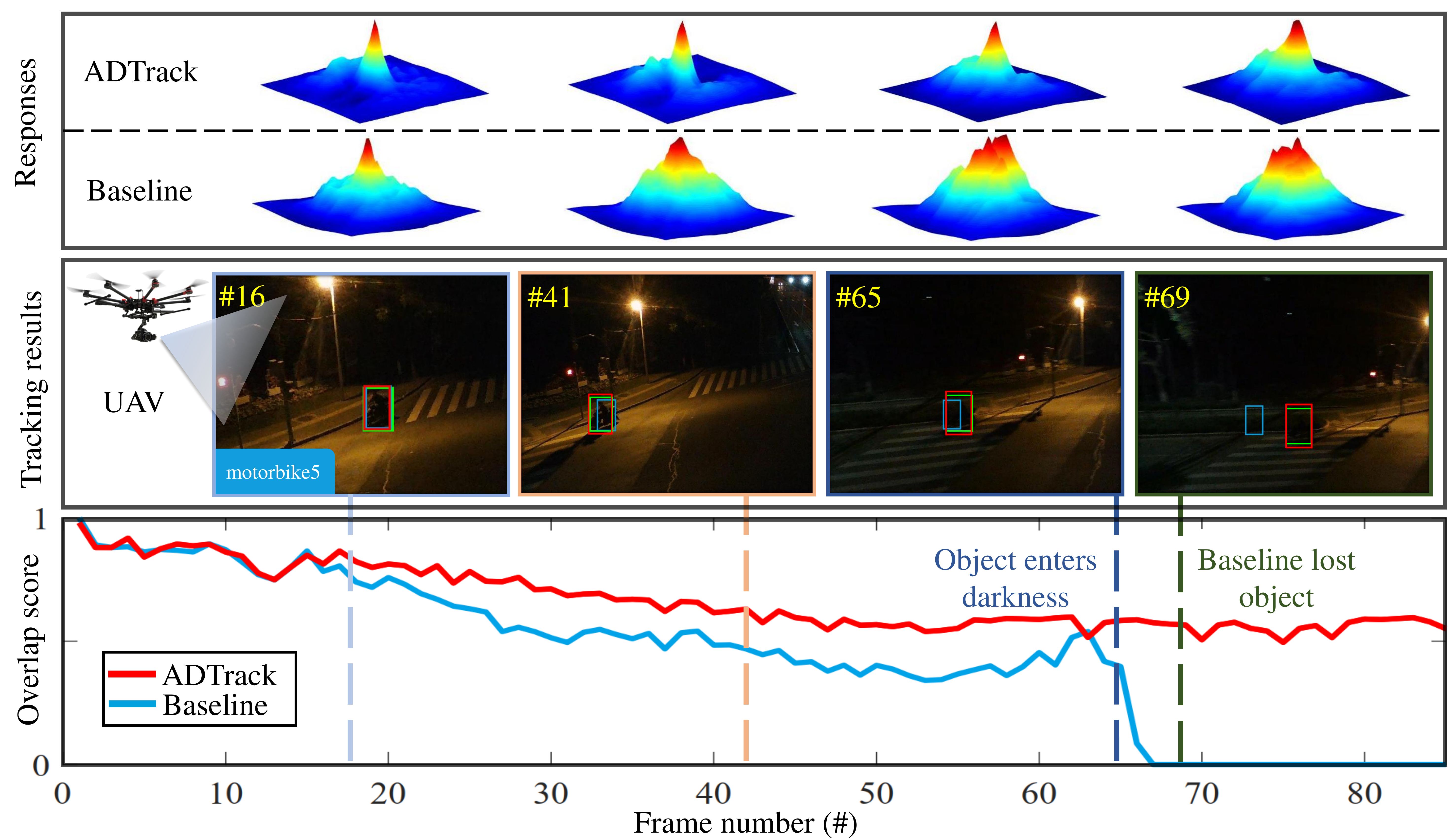}
\caption{Performance comparison of baseline tracker BACF \cite{Galoogahi2017ICCV} and proposed tracker ADTrack in dark sequence \textit{motorbike5}. The shape of response map of BACF tracker is not ideal, which easily leads to tracking failure when the object merges into the dark. While the proposed ADTrack can maintain satisfactory tracking even when the object is invisible in the dark. \textcolor{green}{Green} boxes denote the ground-truth. Some typical dark tracking scenes and performances of the SOTA trackers can be found at \url{https://youtu.be/8ZnGOwoqDZ8}.}
\label{fig:1}
\vspace{-0.3cm}
\end{figure}

Prior work gives few regards to robust tracking in the dark, which is essential and crucial to broaden the application range and service life of UAV. A direct strategy is to couple SOTA low-light enhancement methods \cite{Zhang2019ACM,Park2018Access,Li2018TIP} and trackers~\cite{Bertinetto2016ECCV,Galoogahi2017ICCV,Li2020CVPR}, \textit{i.e.}, operating tracking onto the enhanced images. Even if effective, such a simple fashion has obvious drawbacks: $a$) most SOTA low-light enhancing methods are time-consuming, thereby adds a heavy burden to the overall algorithm; $b$) merely employing preprocessing images for tracking does not fully explore the potential of enhancers; $c$) enhanced images usually have extra noises, interfering the tracker.

To this end, we propose a novel tracker with \textit{Anti-Dark} function (ADTrack), conducting to render real-time and robust tracking onboard UAV at night. To be specific, ADTrack embeds a high-speed low-light image enhancing algorithm into an effective CF-based tracker framework. To our excitement, the image enhancing algorithm can be explored to further generate a target-aware mask based on the illumination information of an image. With the mask, ADTrack proposes a dual regression, where context filter and target-focused filter mutually restrict each other during training, while in the detection stage, the dual filters complement each other for more precise localization. Moreover, the mask favorably filters out the noise brought by the enhancer. Therefore, the proposed ADTrack can maintain splendid tracking performance at night while ensuring real-time tracking speed. Fig.~\ref{fig:1} displays the performance comparison of baseline CF-based tracker~\cite{Galoogahi2017ICCV} and our proposed ADTrack in dark scenes.

In addition, to the best of our knowledge, there exists no dark tracking benchmark now. Hence, this work presents a pioneering UAV dark tracking benchmark (UAVDark70), including 70 videos with a variety of objects. All the HD videos were captured by commercial UAV at night. Contributions\footnote{The source code of the proposed tracker and newly constructed benchmark UAVDark70 are located at \url{https://github.com/vision4robotics/ADTrack}.} of this work can be summarized as:
\begin{itemize}
	\item This work proposes a novel anti-dark tracker, which unites the first stage of an image enhancement methods into CF structure for real-time UAV tracking at night.
	\item This work exploits image illumination variance information to obtain an innovative and effective mask that enables dual regression for dual filter learning and filters out noises, bringing CF-based trackers up to a new level.
	\item Extensive experiments are undertaken on the newly constructed UAVDark70 and UAVDark to demonstrate the robustness and efficiency of ADTrack in the dark.
\end{itemize}

\section{Related Works}
\subsection{Low-Light Enhancement Methods}
Low-light image enhancement algorithms can be generally divided into two categories. The first type like \cite{Zhang2019ACM,Park2018Access}, aims to offline train a deep neural network with numerous pairs of data. The calculation of such methods is too huge to be integrated into UAV real-time trackers. The other is based on retinex theory~\cite{Land1977SA}, without deploying large-scale offline training, \cite{Li2018TIP,Ahn2013ICCE}, which explores illumination and reflectance separated from the whole image to operate them adaptively. In particular, the proposed global adaptation output in \cite{Ahn2013ICCE} is proved to be efficient and effective in low-light enhancement by experiments, which is suitable for integration into UAV tracking algorithm. In addition, the global adaptation output can be further deployed to generate a target-aware mask in this work to elevate robustness.
\subsection{CF-Based Tracking Approaches}
Among various tracking methods, CF-based trackers \cite{Bolme2010CVPR,Henriques2015TPAMI,Danelljan2015ICCV} are popular mainly relying on: $a$) their fast element-wise product in Fourier domain, $b$) online trained filters which are favorably adaptive when object appearance undergoes abrupt variation. On account of the high robustness, adaptability, and efficiency in tracking, CF-based trackers have flourished recently in the field of visual tracking \cite{Bolme2010CVPR,Henriques2015TPAMI,Danelljan2015ICCV,Galoogahi2017ICCV,danelljan2017TPAMI,Danelljan2017CVPR}. Further, CF-based trackers have been demonstrated to be the promising choice for UAV tracking due to their efficiency \cite{Li2020CVPR,Fu2020TGRS,Fu2020GRSM,Fu_2020_TGRS,Ding2020IROS}, where the real-time processing speed on a single CPU platform is crucial.

To be specific, D. S. Bolme \textit{et al.} \cite{Bolme2010CVPR} proposed a seminar MOSSE method as the earliest CF-based tracker. J. F. Henriques \textit{et al.} \cite{Henriques2015TPAMI} introduced kernel function and Tikhonov regularization term for more robust CF-based tracking. H. K. Galoogahi \textit{et al.} \cite{Galoogahi2017ICCV} integrated a cropping matrix during filter training and used alternating direction method of multipliers (ADMM) for optimization, making their tracker aware of the real background information. Oriented at more challenging UAV tracking, \cite{Li2020CVPR} proposed more adaptive and robust AutoTrack with automatic spatio-temporal regularization. While the SOTA CF-based trackers generally perform well during daytime, ignoring robust tracking in the dark. 
\subsection{Target-Aware CF-Based Tracking}
Target-aware mask aims to highlight important parts within target region for filter training. In \cite{Danelljan2015ICCV}, M. Danelljan introduced a fixed spatial penalty, focusing the attention of the filter on learning the center region of extracted samples at a coarse level. A. Lukezic \textit{et al.}~\cite{Lukezic2017CVPR} proposed an adaptive spatial reliability mask based on the Bayes rule. Lately, C. Fu \textit{et al.}~\cite{Fu2020TGRS} employs an efficient saliency detection algorithm to generate an effective mask.

A huge drawback of the prior work is that when an invalid or unreliable mask is generated, wrong regions in the filter will hold higher weights, causing inferior robust tracking or even tracking failure. Besides, images captured in the dark generally possess inadequate information to generate masks in \cite{Fu2020TGRS}, \cite{Lukezic2017CVPR}. To this end, apart from the prior work, ADTrack employs the generated mask from the illumination map to train a dual target-focused filter for restraining the original context filter, which proves to be more robust.
\section{Methodology}
\begin{figure*}[!t]
	\centering
	\setlength{\belowcaptionskip}{-3cm}
	\includegraphics[width=2.04\columnwidth]{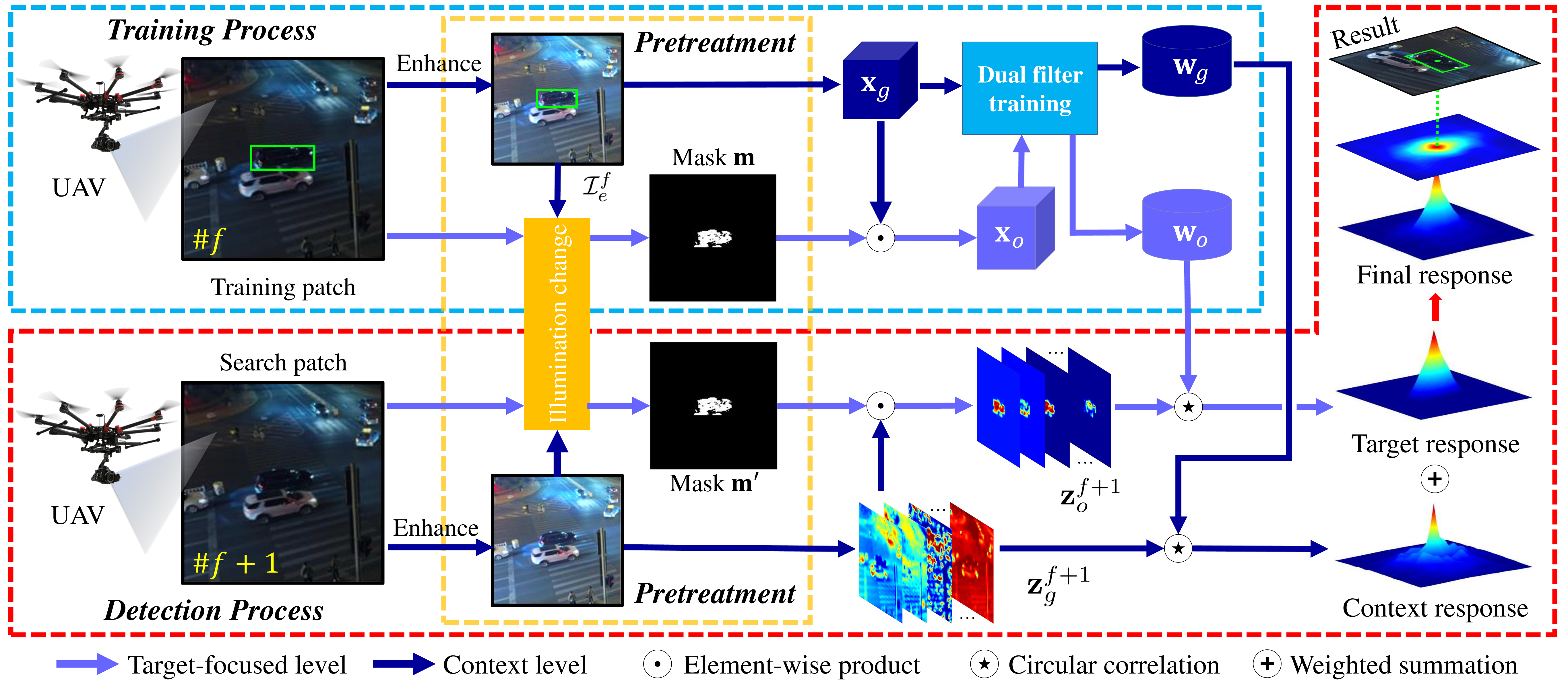}
	\caption{Overall framework of the proposed ADTrack. ADTrack includes 3 stages: pretreatment, training, and detection, which are marked out by boxes in different colors. Dual filters, \textit{i.e.}, context filter and target-focused filter, training and detection follow routes in different colors. It can be seen that the final response shaded noises in context response, which indicates the validity of proposed dual filter. }
	\label{fig:main}
	\vspace{-0.5cm}
\end{figure*}
The pipeline of ADTrack consists of three progressive stages: cropped patch pretreatment, dual filter training and weighted response generation. As shown in Fig.~\ref{fig:main}, when the UAV camera captures the $f$-th frame in the dark, ADTrack firstly implements image pretreatment to achieve image enhancement and mask generation. Then, dual filters are jointly trained by focusing on both context and target appearance. As next frame comes, the trained filters generate dual response maps which are fused to obtain the final response for target localization.
\vspace{-0.1cm}
\subsection{Pretreatment Stage}
As a bio-inspired technique from human front-end visual perception system, ADTrack firstly deploys the beginning of enhancer in \cite{Ahn2013ICCE} to enhance low-light images. When a low-light image RGB $\mathcal{I}\in\mathbb{R}^{w\times h\times 3}$ is input, the pixel-level world illumination value $\mathcal{L}^{\mathrm{W}}(x,y,\mathcal{I})$ is firstly computed as:
{\setlength\abovedisplayskip{6pt}
\setlength\belowdisplayskip{6pt}
\begin{equation}\footnotesize\label{eqn:1}
\mathcal{L}^{\mathrm{W}}(x,y,\mathcal{I})=\sum_{\mathrm{m}}\alpha_{\mathrm{m}}\Psi_{\mathrm{m}}(\mathcal{I}(x,y)),~\mathrm{m}\in\{\mathrm{R,G,B}\}~,
\end{equation}}\noindent where $\Psi_{\mathrm{m}}(\mathcal{I}(x,y))$ indicates the pixel value of image $\mathcal{I}$ at location $(x,y)$ in channel $\mathrm{m}$, \textit{e.g.,} $\Psi_{\mathrm{R}}(\mathcal{I}(x,y))$ denotes the value in red channel. The channel weight parameters $\alpha_\mathrm{R},\alpha_\mathrm{G},\alpha_\mathrm{B}$ meet $\alpha_\mathrm{R}+\alpha_\mathrm{G}+\alpha_\mathrm{B}=1$. Then, the log-average luminance $\tilde{\mathcal{L}}^{W}(\mathcal{I})$ is given as in \cite{Reinhard2002ACM}:
{\setlength\abovedisplayskip{6pt}
\setlength\belowdisplayskip{6pt}
\begin{equation}\footnotesize
\tilde{\mathcal{L}}^{\mathrm{W}}(\mathcal{I})={\mathrm{exp}}\Big(\frac{1}{wh}\sum_{x,y}\mathrm{log}(\delta+\mathcal{L}^{\mathrm{W}}(x,y,\mathcal{I}))\Big)~,
\end{equation}}where $\delta$ is a small value, to avoid zero value. Lastly, the global adaptation factor $\mathcal{L}_{\mathrm{g}}(x,y,\mathcal{I})$ is defined as in \cite{Ahn2013ICCE}:
{\setlength\abovedisplayskip{7pt}
\setlength\belowdisplayskip{7pt}
\begin{equation}\footnotesize \label{eqn:3}
\mathcal{L}_{\mathrm{g}}(x,y,\mathcal{I})=\frac{\mathrm{log}(\mathcal{L}^{\mathrm{W}}(x,y,\mathcal{I})/\tilde{\mathcal{L}}^{\mathrm{W}}(\mathcal{I})+1)}{\mathrm{log}(\mathcal{L}^{\mathrm{W}}_{\mathrm{max}}(\mathcal{I})/\tilde{\mathcal{L}}^{\mathrm{W}}(\mathcal{I})+1)}~,
\end{equation}}where $\mathcal{L}^{\mathrm{W}}_{ \mathrm{max}}(\mathcal{I})=\mathrm{\mathrm{max}}(\mathcal{L}^{\mathrm{W}}(x,y,\mathcal{I}))$. The calculated factor can be referred to change the pixel value in three intensity channels of each pixel to realize image enhancement as:
{\setlength\abovedisplayskip{6pt}
\setlength\belowdisplayskip{6pt}
\begin{equation}\footnotesize \label{eqn:19}
\Psi_{\mathrm{m}}(\mathcal{I}_\mathrm{e}(x,y))=\Psi_{\mathrm{m}}(\mathcal{I}(x,y))\cdot\frac{\mathcal{L}_{\mathrm{g}}(x,y,\mathcal{I})}{\mathcal{L}^{\mathrm{W}}(x,y,\mathcal{I})}~,
\end{equation}}where $\mathcal{I}_{\mathrm{e}}$ denotes the enhanced image based on original $\mathcal{I}$. \\
To our excitement, the algorithm can be used to generate a target-focused mask.
By simple deviation, the illuminance change $\bm{\Theta}_{\mathcal{L}}(\mathcal{I})$ after enhancement can be written as:
{\setlength\abovedisplayskip{6pt}
\setlength\belowdisplayskip{8pt}
\begin{equation}\footnotesize \label{eqn:4}
\bm{\Theta}_{\mathcal{L}}(\mathcal{I})=\mathcal{L}^{\mathrm{W}}(\mathcal{I})-\mathcal{L}^{\mathrm{W}}(\mathcal{I}_\mathrm{e})=\frac{\mathcal{L}^{\mathrm{W}}(x,y,\mathcal{I})-\mathrm{log}\Big(\frac{\mathcal{L}^{\mathrm{W}}(x,y,\mathcal{I})}{\tilde{\mathcal{L}}^{\mathrm{W}}(\mathcal{I})+1}\Big)}{\mathrm{log}(\mathcal{L}^{\mathrm{W}}_{ \mathrm{max}}(\mathcal{I})/\tilde{\mathcal{L}}^{\mathrm{w}}(\mathcal{I})+1)}~.
\end{equation}}
\vspace*{-0.3cm}\\
Since $\mathcal{L}^{\mathrm{W}}(x,y,\mathcal{I})\in[0,1]$, the value of $\bm{\Theta}_{\mathcal{L}}(\mathcal{I})$ apparently varies according to the original illumination. Owing to the fact that different objects' illumination are different under similar light condition in an image due to their various reflectivity, the illumination change $\bm{\Theta}_{\mathcal{L}}(\mathcal{I})$ of different objects varies even bigger. Thereby, by virtue of Eq.~(\ref{eqn:4}), the class of pixels can be indicated as the target or the context. 
\begin{figure}[!b]
	\centering
	\setlength{\abovecaptionskip}{-0.3cm}
	\includegraphics[width=1\columnwidth]{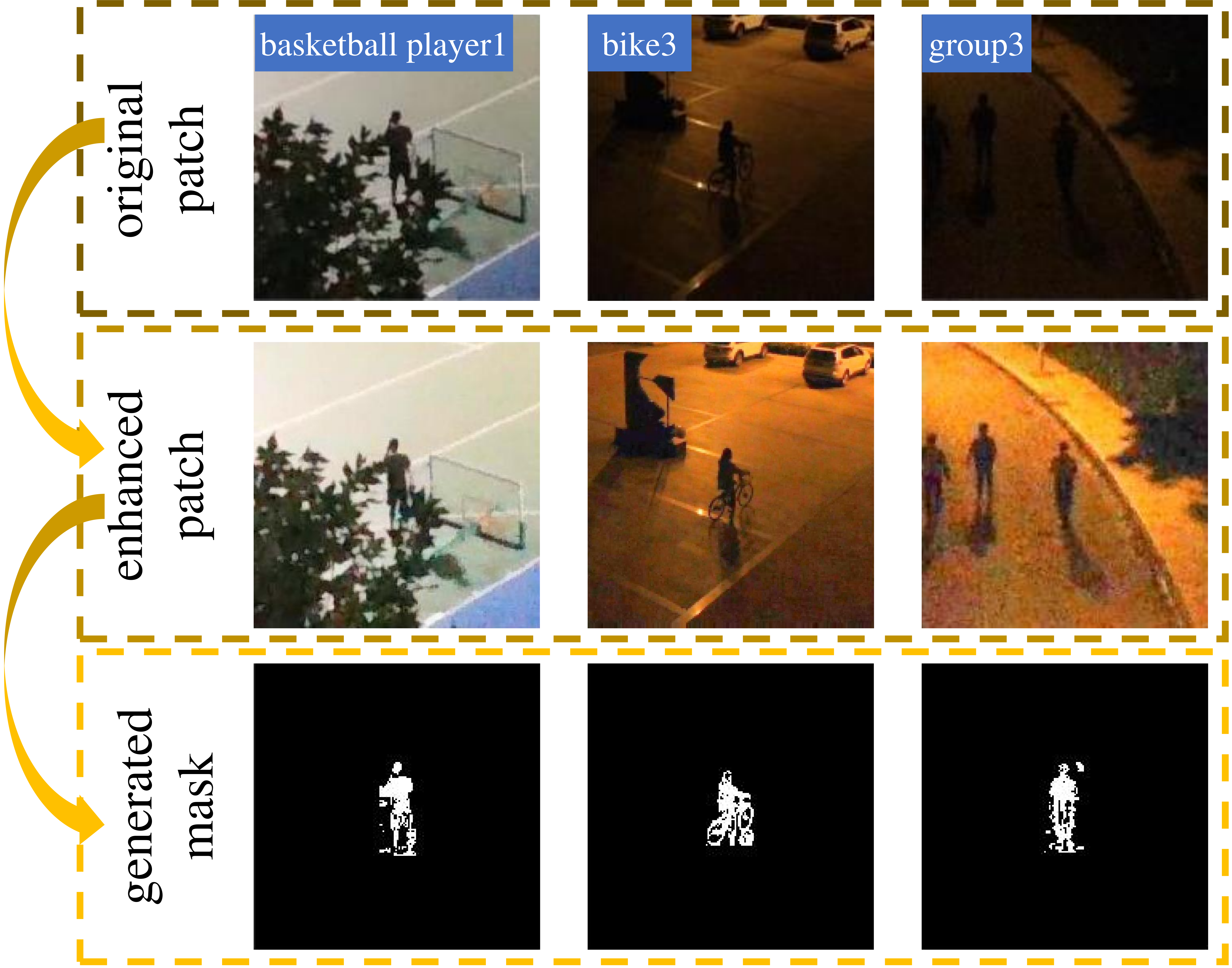}
	\caption{Visualization of pretreatment stage. From top to bottom, the images denote original patch, enhanced patch, and generated mask. The sequences, \textit{i.e.}, \textit{basketball player1}, \textit{bike3}, and \textit{group3}, are from newly constructed UAVDark70. }
	\label{fig:mask}
\end{figure}
To be specific, the average value $\mu$ and standard deviation $\sigma$ of the center region of $\mathbf{\Theta}_{\mathcal{L}}$ are computed. Following a three-sigma criterion in statistics, pixels in the range $\mu\pm3\sigma$ are considered targets while others are the context. Then, a binary mask $\mathbf{m}_r$ is generated, where one is filled into pixels pertaining to segmented target while zero for others. Ultimately, the expected mask is obtained by $\mathbf{m}=\mathbf{m}_r\odot\mathbf{P}$,
where $\odot$ denotes element-wise product. $\mathbf{P}\in\mathbb{R}^{w\times h}$ is the cropping matrix, which extracts the value of the target-size area in the middle of the raw mask $\mathbf{m}_r$, and set the value of the remaining area to 0 to shield the interference of similar brightness objects in the background. 

\noindent\textit{\Remark} Not only can the mask robustly segment object from its background, but it can also block out the noise brought by enhancer, \textit{i.e.}, Eq.(\ref{eqn:4}).
Fig. \ref{fig:mask} exhibits the typical examples of the pretreatment stage.
\vspace{-0.3cm}
\subsection{Training Stage}
\vspace{-0.1cm}
\subsubsection{Review Baseline}
This work adopts background-aware correlation filters (BACF) \cite{Galoogahi2017ICCV} as the baseline tracker due to its outstanding performance stemming from the cropping matrix $\mathbf{P}$. The regression objective to train the BACF is defined as:
{\setlength\abovedisplayskip{9pt}
\setlength\belowdisplayskip{9pt}
\begin{equation}\footnotesize \label{eqn:6}
\mathcal{E}(\mathbf{w})=\frac{1}{2}\sum_{j=1}^{T}\left\|\sum_{c=1}^{D}\mathbf{w}^{c\top}\mathbf{P}\mathbf{C}^j\mathbf{x}^c-\mathbf{y}(j)\right\|_2^2+\frac{\lambda}{2}\sum_{c=1}^{D}\left\|\mathbf{w}^c\right\|_2^2~,
\end{equation}}where $\mathbf{w}^c\in\mathbb{R}^N (c=1,2,\cdots,D)$ is the filter in the $c$-th channel trained in current frame and $\mathbf{w}=[\mathbf{w}^1,\mathbf{w}^2,\cdots,\mathbf{w}^D]$ denotes the whole filter. $\mathbf{x}^c\in\mathbb{R}^T$ is the $c$-th channel of extracted feature map and $\mathbf{y}(j)$ denotes the $j$-th element in the expected Gaussian-shape regression label $\mathbf{y}\in\mathbb{R}^T$. Cropping matrix $\mathbf{P}\in\mathbb{R}^{N\times T}$ aims at cropping the center region of samples $\mathbf{x}^c$ for training and cyclic shift matrix $\mathbf{C}^j\in\mathbb{R}^{T\times T}$ is the same in \cite{Henriques2015TPAMI}, which is employed to obtain cyclic samples. $\lambda$ is the regularization term parameter.

\subsubsection{Training Objective}
Apart from BACF~\cite{Galoogahi2017ICCV}, which trains single filter $\mathbf{w}$ with both negative and positive target-size samples, ADTrack trains dual filters $\mathbf{w}_g$ and $\mathbf{w}_{o}$ by learning context information and target information separately. Besides, a constraint term is added into the overall objective to promise more robust tracking on-the-fly. The proposed regression objective can be written as:
{\setlength\abovedisplayskip{0pt}
	\setlength\belowdisplayskip{6pt}
    \begin{equation}\footnotesize \label{eqn:7}
	\begin{split}
	\mathcal{E}(\mathbf{w}_g,\mathbf{w}_{o})=\sum_{k}\Big(\left\|\sum_{c=1}^{D}\mathbf{P}^{ \top}\mathbf{w}_k^c\star\mathbf{x}_k^c-\mathbf{y}\right\|_2^2
	+\frac{\lambda_1}{2}\sum_{c=1}^{D}\left\|\mathbf{w}_k^c\right\|_2^2\Big)\\
	+\frac{\mu}{2}\sum_{c=1}^{D}\left\|\mathbf{w}_g^c-\mathbf{w}_{o}^c\right\|_2^2~,k\in\{g,o\}~,
	\end{split}
	\end{equation}}where $\star$ denotes circular correlation operator, which implicitly executes sample augmentation by circular shift. Thus the first and third terms formally equivalent to the first term in Eq.~(\ref{eqn:6}). Differently, $\mathbf{x}_g$ denotes the context feature map, while $\mathbf{x}_{o}$ indicates the target region feature map, which is generated using the mask $\mathbf{m}$, \textit{i.e.}, $\mathbf{x}_{o}=\mathbf{m}\odot\mathbf{x}_g$.
	The second and fourth term in Eq.~(\ref{eqn:7}) serve as the regularization term, and the last term can be considered as the constraint term, where $\mathbf{w}_g$ and $\mathbf{w}_{o}$ bind each other during training. $\mu$ is a parameter used to control the impact of the constraint term.
	
\noindent\textit{\Remark}In order to maintain historic appearance information of object, this work follows a conventional fashion in \cite{Galoogahi2017ICCV} for adaptive feature updates using linear interpolation strategy with a fixed learning rate.
\subsubsection{Optimization}
Suppose that $\mathbf{w}_{o}$ is given, ADTrack firstly finds the optimal solution of $\mathbf{w}_g$. Defining an auxiliary variable $\mathbf{v}$, \textit{i.e.}, $\mathbf{v}=\mathbf{I}_N\otimes\mathbf{P}^{\top}\mathbf{w}_g\in\mathbb{R}^{TD}$,
where $\otimes$ denotes Kronecker product, $\mathbf{I}_N$ an $N$-order identical matrix. Here, $\mathbf{w}_g=[\mathbf{w}^{1\top}_g,\mathbf{w}^{2 \top}_g,\cdots,\mathbf{w}^{D\top}_g]^{\top}\in\mathbb{R}^{ND}$. Then, the augmented Lagrangian form of Eq.~(\ref{eqn:7}) is formulated by:
{\setlength\abovedisplayskip{6pt}
\setlength\belowdisplayskip{6pt}
\begin{equation}\footnotesize \label{eqn:10}
\begin{split}
\mathcal{E}(\mathbf{w}_g,\mathbf{v},\bm{\theta})=\frac{1}{2}\left\|\mathbf{v}\star\mathbf{x}-\mathbf{y}\right\|^2_2+\frac{\lambda_1}{2}\left\|\mathbf{w}_g\right\|_2^2+\frac{\mu}{2}\left\|\mathbf{w}_g-\mathbf{w}_{o}\right\|_2^2\\
+(\mathbf{I}_N\otimes\mathbf{P}^{\top}\mathbf{w}_g-\mathbf{v})^{\top}\bm{\theta}+\frac{\gamma}{2}\left\|\mathbf{I}_N\otimes\mathbf{P}^{\top}\mathbf{w}_g-\mathbf{v}\right\|^2_2~,
\end{split}
\end{equation}}where $\bm{\theta}=[\bm{\theta}^{1\top},\bm{\theta}^{2\top},\cdots,\bm{\theta}^{D \top}]^{\top}\in\mathbb{R}^{TD}$ is the Lagrangian vector and $\gamma$ denotes a penalty factor. Then ADMM \cite{Stephen2010FTML} is utilized to iteratively solve $\mathbf{w}_g,\mathbf{v},$ and $\bm{\theta}$.\\
\textbf{Subproblem $\mathbf{w}'_g$}:
By setting the partial derivative of Eq.~(\ref{eqn:10}) with respect to $\mathbf{w}_g$ as zero, we can find the closed-form solution $\mathbf{w}'_g$, which is expressed as:
{\setlength\abovedisplayskip{6pt}
\setlength\belowdisplayskip{6pt}
\begin{equation}\footnotesize \label{eqn:11}
\mathbf{w}'_g=\frac{\mu\mathbf{w}_{o}+T\bm{\theta}+\gamma T \mathbf{v}}{\lambda_1+\mu+\gamma T}~.
\end{equation}}\textbf{Subproblem $\mathbf{v}'$}:
To effectively achieve the closed-form of $\mathbf{v}$, this work firstly turn Eq.~(\ref{eqn:10}) into Fourier domain using discrete Fourier transform as:
{\setlength\abovedisplayskip{9pt}
\setlength\belowdisplayskip{9pt}
\begin{equation}\footnotesize \label{eqn:12}
\begin{split}
\mathbf{v}'=\rm{arg}\min_{\hat{\mathbf{v}}}&\Big\{\frac{1}{2T}\left\|\hat{\mathbf{v}}^{*}\odot\hat{\mathbf{x}}-\hat{\mathbf{y}}\right\|^2_2+\hat{\bm{\theta}}^{\top}(\sqrt{T}\mathbf{I}_N\otimes\mathbf{P}^{\top}\mathbf{F}_N\mathbf{w}_g\\
&-\hat{\mathbf{v}})+\frac{\gamma}{2T}\left\|\sqrt{T}\mathbf{I}_N\otimes\mathbf{P}^{\top}\mathbf{F}_N\mathbf{w}_g-\hat{\mathbf{v}}\right\|^2_2\Big\}~,\\
\end{split}
\end{equation}}where $\hat{\cdot}$ denotes the Fourier form of a variable, \textit{i.e.,} $\hat{\mathbf{x}}=\sqrt{T}\mathbf{F}_T\mathbf{x}$. $\mathbf{F}_T\in\mathbb{C}^{T\times T}$ is the Fourier matrix. Superscript $\cdot^{*}$ indicates the complex conjugate. 

\noindent\textit{\Remark}Since circular correlation is turned into element-wise product in Eq.~(\ref{eqn:12}), by separating sample across pixels, \textit{e.g.,} $\mathbf{x}(t)=[\mathbf{x}^1(t),\mathbf{x}^2(t),\cdots,\mathbf{x}^D(t)]\in\mathbb{R}^{D} (t=1,2,\cdots,T)$, each $\hat{\mathbf{v}}'(t)$ can be solved as:
{\setlength\abovedisplayskip{6pt}
\setlength\belowdisplayskip{6pt}
\begin{equation}\footnotesize \label{eqn:13}
\begin{split}
\hat{\mathbf{v}}'(t)=\Big(\hat{\mathbf{x}}(t)\hat{\mathbf{x}}(t)^{\top}+T\gamma\mathbf{I}_D\Big)^{-1}\Big(\hat{\mathbf{y}}(t)\hat{\mathbf{x}}(t)-T\hat{\bm{\theta}}(t)+T\gamma\hat{\mathbf{w}}_g(t)\Big)~.
\end{split}
\end{equation}}

Then Sherman-Morrison formula \cite{sherman1950AMS} is applied to avoid the inverse operation and Eq.~(\ref{eqn:13}) is turned into: 
{\setlength\abovedisplayskip{6pt}
\setlength\belowdisplayskip{6pt}
\begin{equation}\footnotesize \label{eqn:14}
\begin{split}
\hat{\mathbf{v}}'(t)=\frac{1}{\gamma T}\Big(\hat{\mathbf{y}}(t)\hat{\mathbf{x}}(t)-T\hat{\bm{\theta}}(t)+\gamma T\hat{\mathbf{w}}_g(t)\Big)-\\
\frac{\hat{\mathbf{x}}(t)}{\gamma Tb}\Big(\hat{\mathbf{y}}(t)\hat{\mathbf{s}}_{\mathbf{x}}(t)-T\hat{\mathbf{s}}_{\bm{\theta}}(t)+\gamma T\hat{\mathbf{s}}_{\bm{w}_g}(t)\Big)~,
\end{split}
\end{equation}}where $\hat{\mathbf{s}}_{\mathbf{x}}(t)=\hat{\mathbf{x}}(t)^{\top}\hat{\mathbf{x}}(t),\hat{\mathbf{s}}_{\bm{\theta}}=\hat{\mathbf{x}}(t)^{\top}\hat{\mathbf{\theta}},\hat{\mathbf{s}}_{\bm{w}_g}=\hat{\mathbf{x}}(t)^{\top}\hat{\mathbf{w}}_g $ and $b=\hat{\mathbf{s}}_{\mathbf{x}}(t)+T\gamma$ are scalar.\\
\textbf{Lagrangian Update}: Having solved $\mathbf{v}$ and $\mathbf{w}_g$ in current $e$-th iteration, the Lagrangian multipliers are updated as:
{\setlength\abovedisplayskip{6pt}
\setlength\belowdisplayskip{6pt}
\begin{equation}\footnotesize \label{eqn:15}
\begin{split}
\hat{\bm{\theta}}^e=\hat{\bm{\theta}}^{e-1}+\gamma(\hat{\mathbf{v}}^e-(\mathbf{FP}^{\top}\otimes\mathbf{I}_{D})\mathbf{w}^e_g)~,
\end{split}
\end{equation}}where the superscript $\cdot^{e}$ indicates current $e$-th iteration.

\noindent\textit{\Remark} The positions of $\mathbf{w}_g$ and $\mathbf{w}_{o}$ in Eq.~(\ref{eqn:7}) are equivalent. When an solving iteration of $\mathbf{w}_g$ is completed, then the same ADMM iteration operation is performed to obtain the optimized solution of $\mathbf{w}_{o}$. 
\vspace{-0.3cm}
\subsection{Detection Stage}
Given the expected filter $\mathbf{w}^f_g$ and $\mathbf{w}^f_{o}$ in the $f$-th frame, the response map $\mathbf{R}$ regarding the detection samples $\mathbf{z}^{f+1}$ in the $(f+1)$-th frame can be obtained by:
{\setlength\abovedisplayskip{6pt}
\setlength\belowdisplayskip{6pt}
\begin{equation}\footnotesize
\label{eqn:16}
\begin{split}
\mathbf{R}=\mathcal{F}^{-1}\sum_{c=1}^D\big(\hat{\mathbf{w}}^{f,c*}_g\odot\hat{\mathbf{z}}^{f+1,c}_g+\psi\hat{\mathbf{w}}^{f,c*}_{o}\odot\hat{\mathbf{z}}^{f+1,c}_{o}\big)~,
\end{split}
\end{equation}}where $\mathcal{F}^{-1}$ means inverse Fourier transform. $\mathbf{z}_g^{f+1,c}$ denotes the $c$-th channel of resized search region samples extracted in the $(f+1)$-th frame, and $\mathbf{z}_{o}^{f+1,c}$ is the $c$-th channel of the masked samples similar to $\mathbf{x}_{o}$. $\psi$ is a weight parameter that controls the impact response map generated by context filter and object filter. Finally, the object location in the $(f+1)$-th frame can be found at the peak of response map $\mathbf{R}$.
\section{Experiment}
\begin{figure}[!t]
	\begin{center}
		\setlength{\abovecaptionskip}{-1cm}
		\setlength{\belowcaptionskip}{-20cm}
		\subfigure{\label{fig:UAVDark70_Pre} 
			\begin{minipage}{0.22\textwidth}
				\centering
				\includegraphics[width=1\columnwidth]{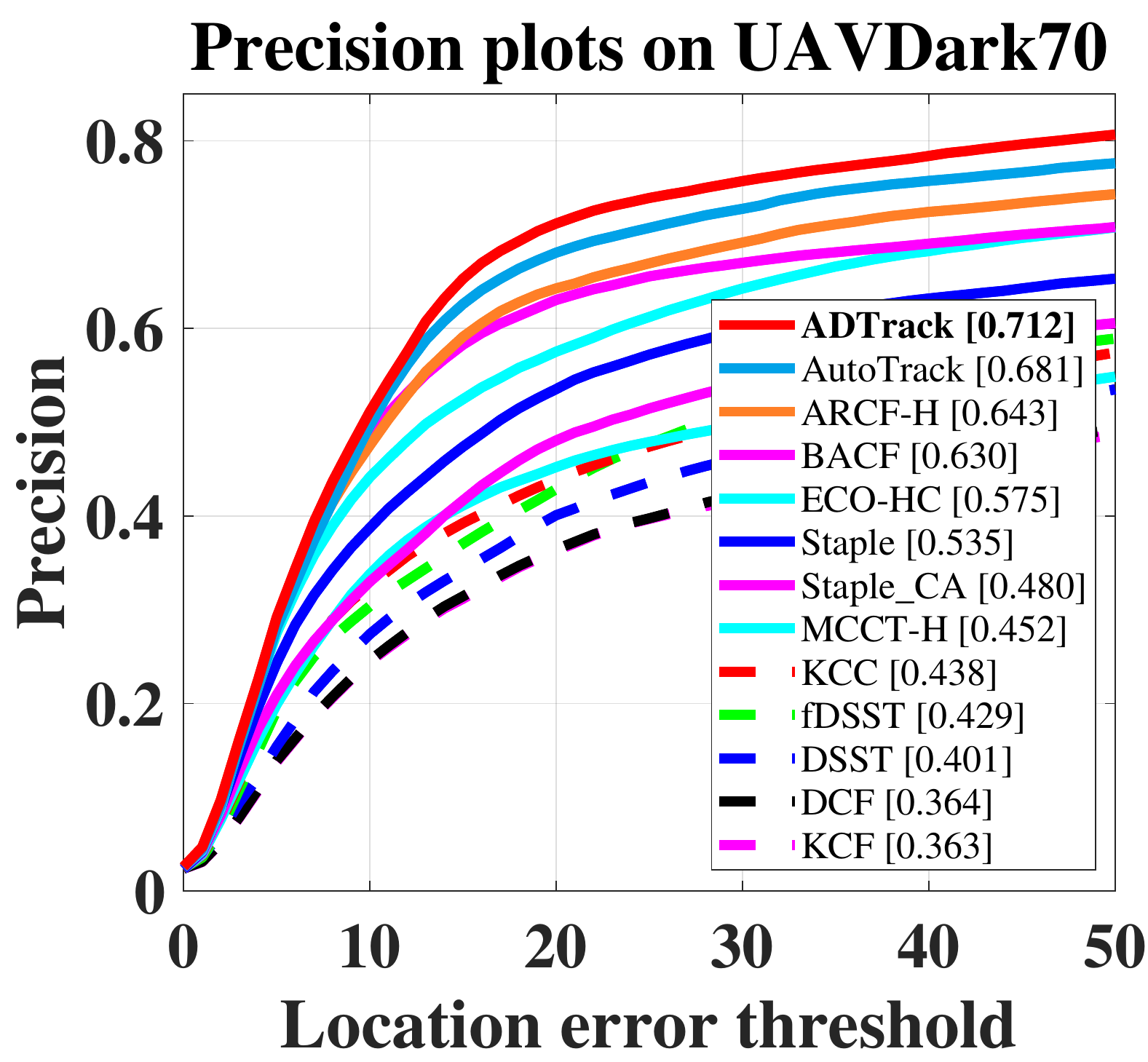}
			\end{minipage}
		}
		\subfigure{\label{fig:UAVDark70_Suc} 
			\begin{minipage}{0.22\textwidth}
				\centering
				\includegraphics[width=1\columnwidth]{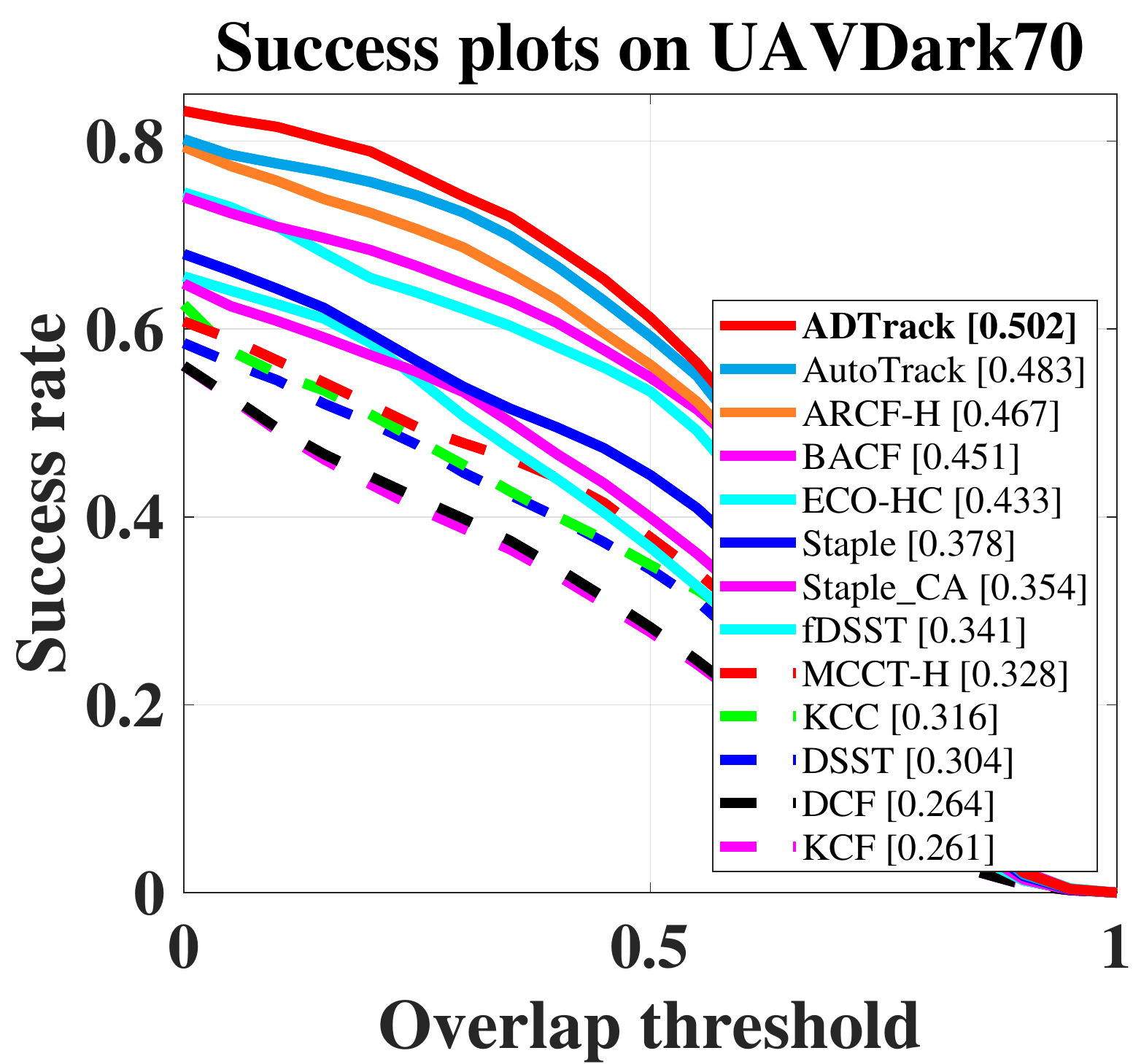}
			\end{minipage}
		}
	\vspace{-0.3cm}
	\begin{center}\footnotesize
		($a$) Results on UAVDark70
	\end{center}
	\vspace{-0.2cm}
		\subfigure{\label{fig:UAVDark_Pre} 
			\begin{minipage}{0.22\textwidth}
				\centering
				\includegraphics[width=1\columnwidth]{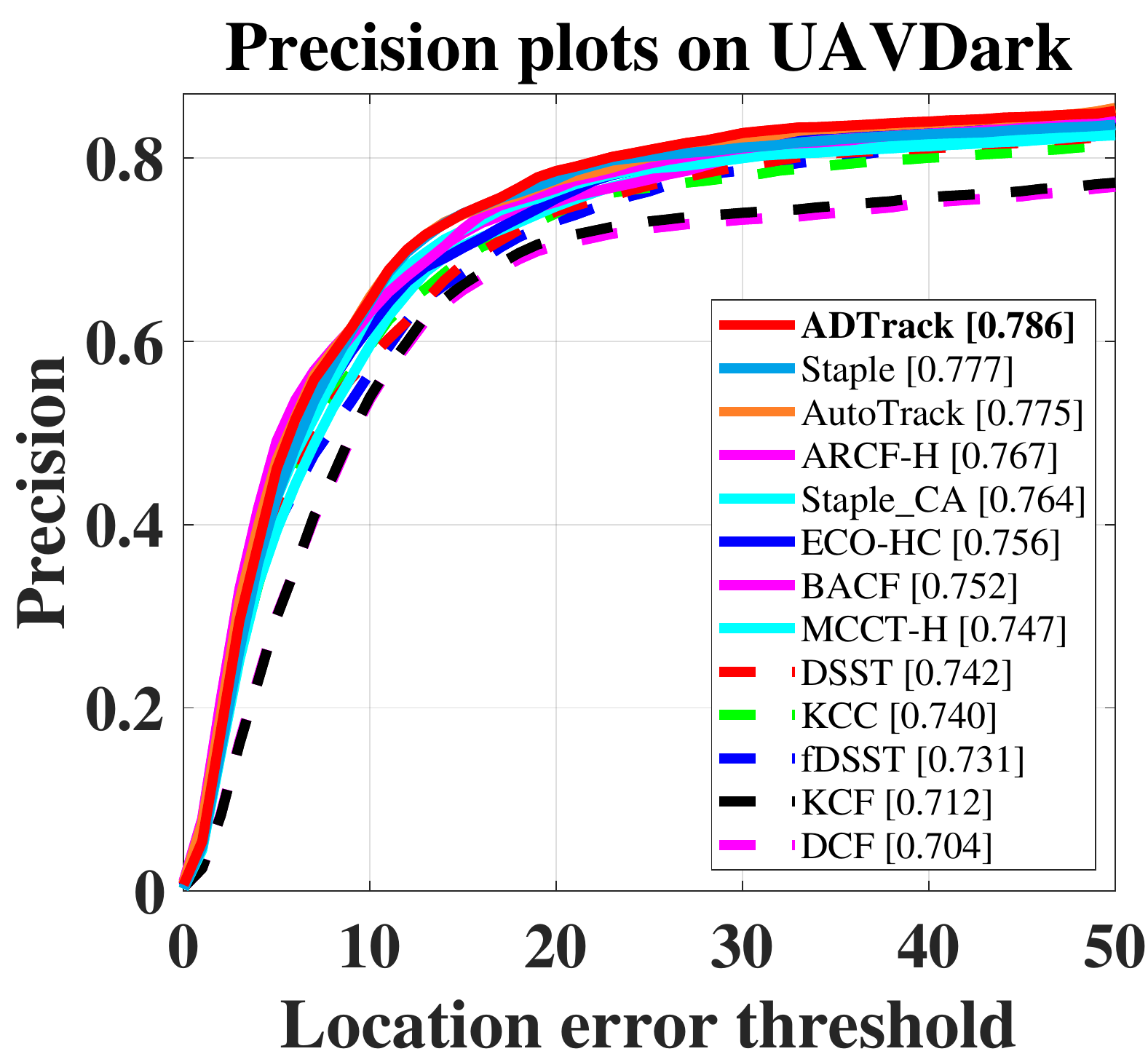}
			\end{minipage}
		}
		\subfigure{\label{fig:UAVDark_Suc} 
			\begin{minipage}{0.22\textwidth}
				\centering
				\includegraphics[width=1\columnwidth]{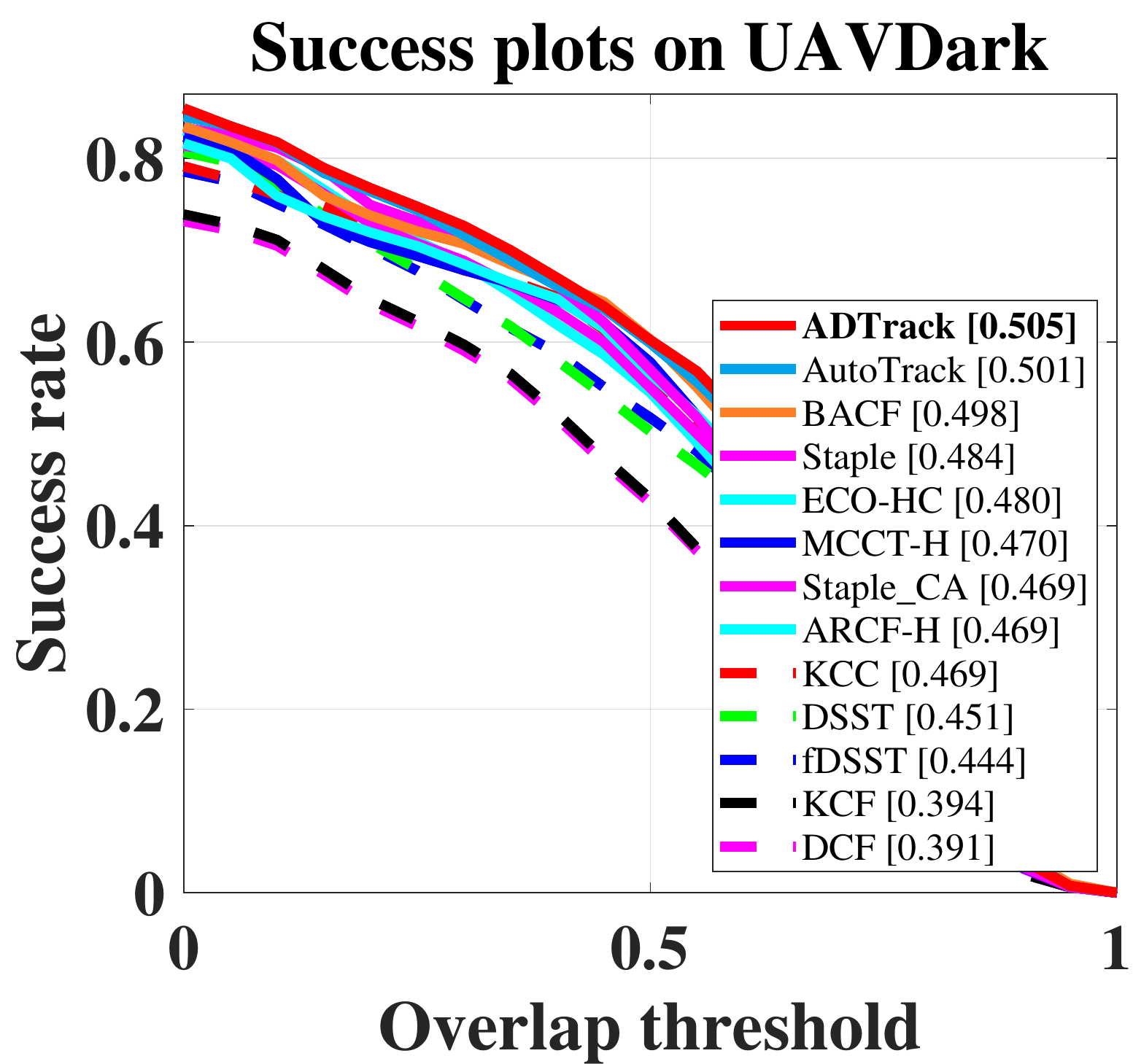}
			\end{minipage}
		}
		\vspace{-0.3cm}
		\begin{center}\footnotesize
		($b$) Results on UAVDark
		\end{center}
	\vspace{-0.6cm}
	\end{center}
	\caption{Overall performance of real-time hand-crafted DCF-based trackers on the benchmark UAVDark70 and UAVDark. The evaluation metric in precision plot is DP, and the metric in success rate plot is AUC. Clearly ADTrack maintains its robustness in 2 benchmarks by virtue of its dual regression.}
	\label{fig:overall}
	\vspace{-0.3cm}
\end{figure}

This part exhibits the exhaustive experimental results. Generally, in Section \ref{Overall}, 16 SOTA hand-crafted CF-based trackers, \textit{i.e.}, AutoTrack \cite{Li2020CVPR}, KCF \& DCF \cite{Henriques2015TPAMI}, SRDCF \cite{Danelljan2015ICCV}, STRCF \cite{Li2018CVPR}, BACF \cite{Galoogahi2017ICCV}, DSST \& fDSST \cite{danelljan2017TPAMI}, ECO-HC \cite{Danelljan2017CVPR}, ARCF-HC \& ARCF-H \cite{Huang2019ICCV}, KCC \cite{wang2018AAAI}, MCCT-H \cite{wang2018CVPR}, CSR-DCF \cite{Lukezic2017CVPR}, Staple \cite{bertinetto2016CVPR}, Staple\_CA \cite{mueller2017CVPR}, and proposed ADTrack are invited for evaluation on two dark tracking benchmark, \textit{i.e.}, UAVDark70 and UAVDark, to demonstrate the superiority of the proposed ADTrack comprehensively. Specially, in Section \ref{deep} as displayed in TABLE~\ref{tab:deep} and TABLE~\ref{tab:deep_ours}, deep trackers deploying convolutional neural network (CNN), have also been evaluated.

\subsection{Implementation Information}

\subsubsection{Platform}
The experiments extended in this work were all performed on MATLAB R2019a. The main hardware adopted consists of an Intel Core I7-8700K CPU, 32GB RAM, and an NVIDIA RTX 2080 GPU.

\subsubsection{Parameters}
To guarantee the fairness and objectivity of the evaluation, the tested trackers from other works have maintained their official initial parameters.

The parameters of the regression equation in ADTrack are as follows, $\lambda_1=\lambda_2=0.01$ and $\mu$ is set as 200. For the hyper parameters of ADMM, ADTrack sets $\gamma_{\rm max}=10000, \gamma^{0}=1$, and the numbers of iteration for $\mathbf{w}_g$ and $\mathbf{w}_o$ are both 3. During detection, weight $\psi$ is set as $0.02$. Note that both context and target appearance adopts a learning rate $\eta_1=0.02$ to implement feature update.

\subsubsection{Features and Scale Estimation}
ADTrack uses hand-crafted features for appearance representations, \textit{i.e.}, gray-scale, a fast version of histogram of oriented gradient (fHOG) \cite{Felzenszwalb2010TPAMI}, and color names (CN) \cite{Weijer2006ECCV}. Note that gray-scale and CN features can be valid in ADTrack thanks to low-light enhancement. The cell size for feature extraction is set as $4\times 4$. ADTrack adopts the scale filter proposed by \cite{danelljan2017TPAMI} to perform accurate scale estimation.
\vspace{-0.1cm}
\begin{figure}[!t]
	\centering
	\setlength{\abovecaptionskip}{-0.3cm}
	\setlength{\belowcaptionskip}{-20cm}
	\includegraphics[width=1\columnwidth]{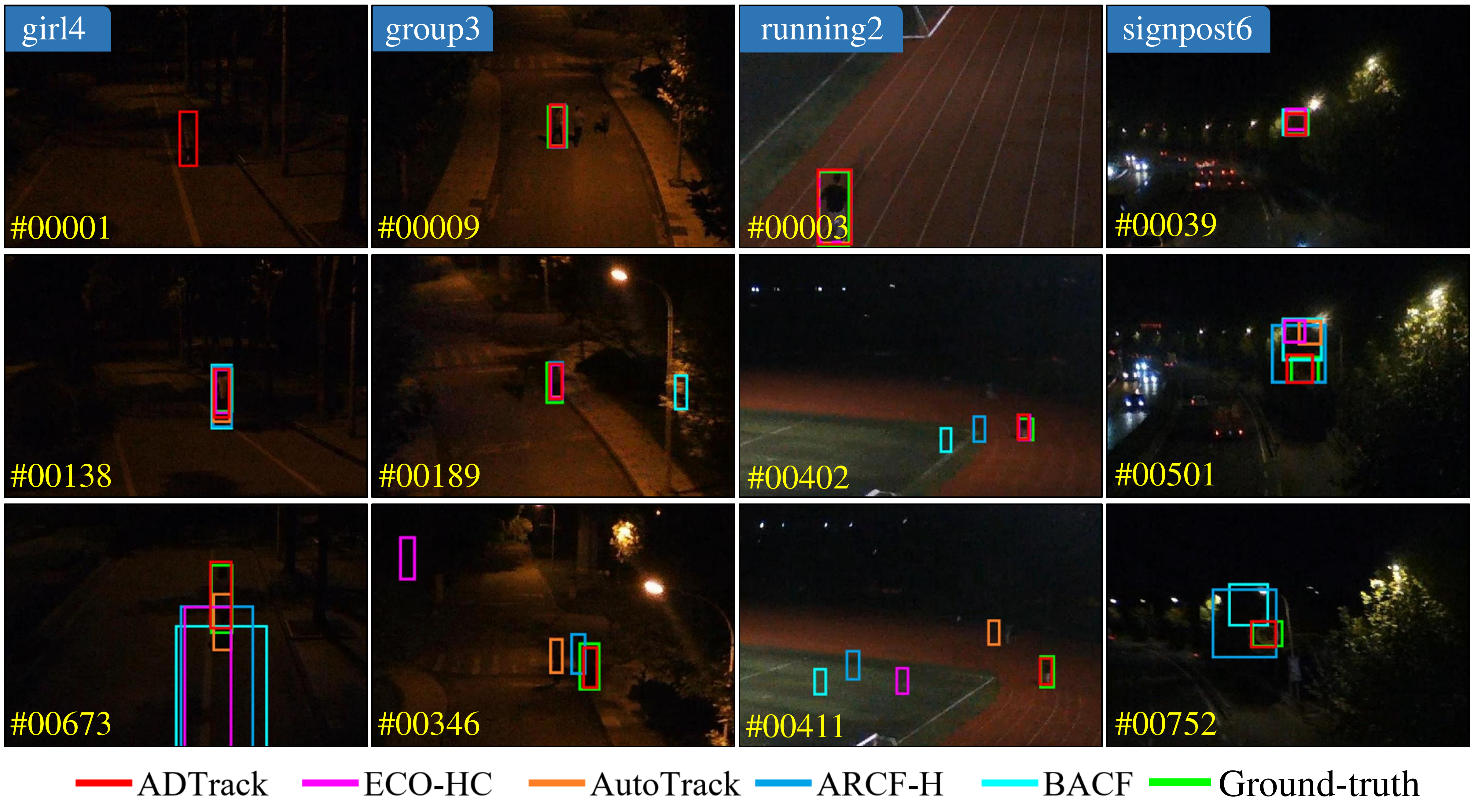}
	\vspace{-0.2cm}
	\caption{Visualization of some typical tracking scenes. Sequences, \textit{i.e.}, \textit{girl4}, \textit{group3}, \textit{running2}, and \textit{signpost6}, which indicate the targets, are from the newly constructed UAVDark70. Clearly, when other trackers lost the object in the dark, ADTrack ensured its robustness in darkness mainly due to its target-aware mask and dual regression. }
	\label{fig:vis}
	\vspace{-0.3cm}
\end{figure}
\subsection{Overall Evaluation}\label{Overall}
Figure~\ref{fig:overall} shows the overall success rate and precision comparison of the real-time trackers.

Benchmark UAVDark70 is newly made in this work, consisting of 70 manually annotated sequences. All the scenes were shot from a professional grade UAV at night. In Fig. \ref{fig:overall}($a$), ADTrack outperforms all other arts and improves the baseline BACF \cite{Galoogahi2017ICCV} tracker by \textbf{10.2\%} under distance precision (DP) at center location error (CLE) = 20 pixels. In terms of area under curve (AUC), ADTrack ranks the first as well. Fig.~\ref{fig:vis} displays some typical dark tracking scenes and performance of the SOTA trackers in UAVDark70.

In order to maintain the objectivity, this work selected typical night scenes from the authoritative publicly available benchmark UAVDT \cite{du2018ECCV} and Visdrone-2019SOT \cite{du2019ICCVW} (totally 37 sequences), and composed them into a new benchmark, \textit{i.e.}, UAVDark. In Fig.~\ref{fig:overall}($b$), ADTrack outperforms all others in both DP and AUC. Specifically, ADTrack improves the DP of baseline BACF by more than \textbf{4.5\%} in UAVDark. 

TABLE~\ref{tab:overa} shows the top 11 hand-crafted CF-based trackers' (both real-time and not real-time) venues and average results on 2 benchmarks, where ADTrack outperforms all other hand-crafted trackers. Besides, ADTrack achieves a speed of 34 FPS, meeting real-time requirement on UAV tracking.

\noindent\textit{\Remark}The sequences in the newly made UAVDark70 are more common in real-world dark tracking, where the scenes are generally much darker, bringing more challenges to trackers.
\vspace{-0.3cm}
\begin{table*}[!t]
	\centering
	\setlength{\tabcolsep}{1.5mm}
	\fontsize{6.5}{8}\selectfont
	\begin{threeparttable}
		\caption{Average results of the selected top 11 trackers using hand-crafted feature. \textbf{\textcolor[rgb]{ 1,  0,  0}{Red}}, \textbf{\textcolor[rgb]{ 0,  1,  0}{green}}, and \textbf{\textcolor[rgb]{ 0,  0,  1}{blue}} respectively mean the first, second and third place. The FPS values in this table are all obtained on a single CPU.}
		\vspace{0.08cm}
		\begin{tabular}{cccccccccccc}
			\toprule[1.5pt]
			Tracker & \textbf{ADTrack} & AutoTrack\cite{Li2020CVPR}& ARCF-HC\cite{Huang2019ICCV} & ARCF-H\cite{Huang2019ICCV} & STRCF\cite{Li2018CVPR} & MCCT-H\cite{wang2018CVPR}&BACF \cite{Galoogahi2017ICCV} & CSR-DCF \cite{Lukezic2017CVPR} & Staple\_CA \cite{mueller2017CVPR} & ECO-HC \cite{Danelljan2017CVPR} & Staple \cite{bertinetto2016CVPR}  \\
			\midrule
			Venue & \textbf{Ours} & '20 CVPR & '19 ICCV & '19 ICCV & '18 CVPR & '18 CVPR & '17 ICCV & '17 CVPR & '17 CVPR & '17 CVPR & '16 CVPR \\
			AUC & \textbf{\textcolor{red}{0.504}}& \textbf{\textcolor{blue}{0.492}} & \textbf{\textcolor{green}{0.497}}& 0.468& 0.492 & 0.399 & 0.484 & 0.428 & 0.412 & 0.457 & 0.431\\
			DP & \textbf{\textcolor{red}{0.749}}&\textbf{\textcolor{green}{0.728}} & \textbf{\textcolor{blue}{0.722}}&0.705& 0.706 & 0.600& 0.699&0.650&0.622&0.666&0.656\\
			FPS & 34.84 & \textbf{\textcolor{blue}{49.05}} & 24.71&38.58 & 22.84 & 47.16& 41.52& 8.42&48.99&\textbf{\textcolor{green}{57.52}}&\textbf{\textcolor{red}{85.16}}\\
			\bottomrule[1.5pt]			
		\end{tabular}\label{tab:overa}
	\end{threeparttable}
	\vspace{-0.5cm}
\end{table*}
\subsection{Analysis by Attributes}
\begin{table}[t]
	\centering
	\setlength{\tabcolsep}{0.6mm}
	\fontsize{5}{8}\selectfont
	\scriptsize
	\begin{threeparttable}
		\caption{Results comparison of the top 8 real-time hand-crafted CF-based trackers on UAVDark70 and UAVDark by UAV-specific attribute. \textbf{\textcolor[rgb]{ 1,  0,  0}{Red}}, \textbf{\textcolor[rgb]{ 0,  1,  0}{green}}, and \textbf{\textcolor[rgb]{ 0,  0,  1}{blue}} respectively mean the first, second and third place. The results here are the average by all sequences involved.}
		\vspace{0.05cm}
		\begin{tabular}{cccccccccccccccccccc}
			\toprule[1.5pt]
			\multirow{2}{*}{\diagbox{Tracker}{Metric}}&
			\multicolumn{5}{c}{AUC}&\multicolumn{5}{c}{DP}\cr
			\cmidrule(lr){2-6} \cmidrule(lr){7-11}
			&IV&OCC&LR&FM&VC&IV&OCC&LR&FM&VC\cr			
			\midrule
			Staple \cite{bertinetto2016CVPR}  &0.436&0.396&0.356&0.433&0.421&0.656&0.561&0.633&0.613&0.632\cr
			ECO-HC \cite{Danelljan2017CVPR} &0.443&0.437&0.363&0.457&0.437&0.580&0.558&0.586&0.620&0.633\cr
			Staple\_CA \cite{mueller2017CVPR}  &0.415&0.410&0.359&0.425&0.396&0.563&0.511&0.622&0.592&0.591\cr
			BACF \cite{Galoogahi2017ICCV} &\textbf{\textcolor{green}{0.482}}&0.444&\textbf{\textcolor{blue}{0.406}}&0.483&\textbf{\textcolor{blue}{0.457}}&0.677&0.607&0.648&0.660&0.663\cr
			MCCT-H \cite{wang2018CVPR} &0.393&0.398&0.322&0.409&0.377&0.591&0.531&0.578&0.567&0.565\cr
			ARCF-H \cite{Huang2019ICCV}&0.465&\textbf{\textcolor{blue}{0.453}}&0.373&\textbf{\textcolor{green}{0.487}}&0.447&\textbf{\textcolor{blue}{0.689}}&\textbf{\textcolor{blue}{0.617}}&\textbf{\textcolor{blue}{0.660}}&\textbf{\textcolor{blue}{0.671}}&\textbf{\textcolor{blue}{0.671}}\cr
			AutoTrack \cite{Li2020CVPR} &\textbf{\textcolor{blue}{0.479}}&\textbf{\textcolor{red}{0.460}}&\textbf{\textcolor{green}{0.441}}&\textbf{\textcolor{blue}{0.487}}&\textbf{\textcolor{green}{0.470}}&\textbf{\textcolor{green}{0.701}}&\textbf{\textcolor{green}{0.638}}&\textbf{\textcolor{green}{0.709}}&\textbf{\textcolor{green}{0.688}}&\textbf{\textcolor{green}{0.698}}\cr
			\textbf{ADTrack} \textbf{(ours)}&\textbf{\textcolor{red}{0.497}}&\textbf{\textcolor{green}{0.459}}&\textbf{\textcolor{red}{0.444}}&\textbf{\textcolor{red}{0.503}}&\textbf{\textcolor{red}{0.485}}&\textbf{\textcolor{red}{0.731}}&\textbf{\textcolor{red}{0.638}}&\textbf{\textcolor{red}{0.720}}&\textbf{\textcolor{red}{0.712}}&\textbf{\textcolor{red}{0.722}}\cr
			\bottomrule[1.5pt]
		\end{tabular}\label{tab:att}
	\end{threeparttable}
\vspace{-0.3cm}
\end{table}
Following \cite{Fu2020GRSM}, this work considers the UAV special tracking challenges as low resolution (LR), fast motion (FM), illumination variation (IV), viewpoint change (VC), and occlusion (OCC). In terms of UAVDark70, the attributes are manually annotated according to the same criterion in \cite{du2019ICCVW}. TABLE~\ref{tab:att} exhibits the average by sequences results of the top 8 real-time CF-based trackers in UAVDark and UAVDark70 according to TABLE~\ref{tab:overa}, where ADTrack ranks the first in most attributes.

\subsection{Ablation Studies}

This part exhibits the validity of different components in ADTrack on tracking. BACF\_e denotes adding only the enhancing factor on the baseline tracker BACF \cite{Galoogahi2017ICCV}. ADTrack\_e means ADTrack without dual filters (considered the Baseline). ADTrack\_{ew} indicates adding merely weighted summation in detection stage on ADTrack\_e. ADTrack means the full version of the proposed tracker (both weighted summation and constraint term). The results are displayed in TABLE~\ref{tab:abla}, where clearly, the proposed 2 components have boosted tracking performance by a large margin, improving more than \textbf{3\%} in both AUC and DP. 

\noindent\textit{\Remark} BACF\_e is worse than original BACF, probably due to the noise introduced by image enhancing. While for ADTrack, the mask can block such negative effect.

\begin{table}[!t]
	\centering
	\setlength{\tabcolsep}{3mm}
	\fontsize{6}{6}\selectfont
	\begin{threeparttable}
		\setlength{\tabcolsep}{3mm}
		\caption{AUC and DP comparison of different versions of ADTrack on UAVDark70. The tracker ADTrack\_e, ADTrack\_{ew} respectively denote ADTrack with different components. }
		\vspace{0.08cm}
		\begin{tabular}{cccccc}
			\toprule[1.5pt]
			Tracker &ADTrack& ADTrack\_{ew} & ADTrack\_{e}& BACF\_{e}& BACF\\
			\midrule
			AUC & \textbf{0.502} &0.492 &0.487&0.448 & 0.451\\
			$\Delta$(\%) & \textbf{+11.3} &+9.0 &+7.9& -0.7 & 0\\
			DP & \textbf{0.712} &0.694 &0.689&0.618 & 0.630\\
			$\Delta$(\%) & \textbf{+13.0} &+10.2 & +9.4& -1.9 & 0\\
			\bottomrule[1.5pt]			
		\end{tabular}\label{tab:abla}
	\end{threeparttable}
\end{table}

\subsection{Against the Deep Trackers}\label{deep}
This section focuses on comparison between proposed ADTrack and deep trackers which utilize off-line trained deep network for feature extraction or template matching. This work invites totally 10 SOTA deep trackers, \textit{i.e.}, SiameseFC \cite{Bertinetto2016ECCV}, ASRCF \cite{Dai2019CVPR}, ECO \cite{Danelljan2017CVPR}, UDT+ \cite{wang2019CVPR}, TADT \cite{li2019CVPR}, UDT \cite{wang2019CVPR}, HCFT \cite{ma2015ICCV}, IBCCF \cite{li2017ICCV}, and DSiam \cite{Guo2017ICCV}, to test their performance in UAVDark. From TABLE~\ref{tab:deep}, ADTrack clearly outperforms most deep trackers in terms of DP and AUC.

\noindent\textit{\Remark}Using merely single CPU, ADTrack still achieves a real-time speed at more than 30 FPS, while many deep trackers are far from real-time even on GPU, demonstrating the excellence of ADTrack for real-time UAV tracking.

TABLE~\ref{tab:deep_ours} selects the top 5 trackers in TABLE~\ref{tab:deep} to evaluate their performance under the newly constructed UAVDark70.

\noindent\textit{\Remark}The results illustrate that the SOTA deep trackers fail to maintain their robustness in real-world common dark scenes, since the CNNs they utilize are trained by daytime images, ending up in their huge inferiority compared with online-learned ADTrack in the dark.
\begin{table}[t]
	\centering
	\setlength{\tabcolsep}{2mm}
	\fontsize{8}{5}\selectfont
	\begin{threeparttable}
		\caption{AUC, DP, and tracking speed (FPS) comparison of the deep trackers and ADTrack on UAVDark. * denotes GPU speed, which is not commonly used in UAV platform. \textbf{\textcolor[rgb]{ 1,  0,  0}{Red}}, \textbf{\textcolor[rgb]{ 0,  1,  0}{green}}, and \textbf{\textcolor[rgb]{ 0,  0,  1}{blue}} respectively mean the first, second and third place.}
		\vspace{0.08cm}
		\begin{tabular}{cccccc}
			\toprule[1.5pt]
			Tracker & Venue & DP& AUC&FPS& GPU\\
			\midrule
			HCFT \cite{ma2015ICCV}& '15 ICCV&0.721&0.451 & 18.26*&\Checkmark\\
			SiameseFC \cite{Bertinetto2016ECCV}& '16 ECCV&  0.713&0.467 & 37.17*&\Checkmark\\
			IBCCF \cite{li2017ICCV}& '17 ICCV&0.731&0.474 & 2.77*&\Checkmark\\
			DSiam \cite{Guo2017ICCV} & '17 ICCV&0.653 & 0.419& 15.62*&\Checkmark\\
			ECO \cite{Danelljan2017CVPR}& '17 CVPR& \textbf{\textcolor{red}{0.790}} & \textbf{\textcolor{blue}{0.498}}& 16.12*&\Checkmark\\
			UDT \cite{wang2019CVPR}& '19 CVPR& 0.754&0.484 & \textbf{\textcolor{red}{56.68*}}&\Checkmark\\
			TADT \cite{li2019CVPR}& '19 CVPR& \textbf{\textcolor{blue}{0.780}}&0.498 & 29.06*&\Checkmark\\
			UDT+ \cite{wang2019CVPR} & '19 CVPR& 0.728&0.459 & \textbf{\textcolor{green}{53.96*}}&\Checkmark \\
			ASRCF \cite{Dai2019CVPR} & '19 CVPR & 0.775 &\textbf{\textcolor{green}{0.500}}& 21.39*&\Checkmark\\
			\textbf{ADTrack} & \textbf{Ours} & \textbf{\textcolor{green}{0.786}}&\textbf{\textcolor{red}{0.505}} &\textbf{\textcolor{blue}{37.71}}&\XSolidBrush\\
			\bottomrule[1.5pt]			
		\end{tabular}\label{tab:deep}
	\end{threeparttable}
\vspace{-0.2cm}
\end{table}

\begin{table}[t]
	\centering
	\setlength{\tabcolsep}{1.2mm}
	\fontsize{8}{8}\selectfont
	\begin{threeparttable}
		\caption{AUC and DP comparison of top 5 trackers in TABLE.~\ref{tab:deep} on UAVDark70. \textbf{\textcolor[rgb]{ 1,  0,  0}{Red}}, \textbf{\textcolor[rgb]{ 0,  1,  0}{green}}, and \textbf{\textcolor[rgb]{ 0,  0,  1}{blue}} respectively mean the first, second and third place.}
		\vspace{0.08cm}
		\begin{tabular}{cccccc}
			\toprule[1.5pt]
			Tracker &\textbf{ADTrack}& ECO \cite{Danelljan2017CVPR} & TADT \cite{li2019CVPR} & ASRCF \cite{Dai2019CVPR} & UDT \cite{wang2019CVPR} \\
			\midrule
			AUC & \textbf{\textcolor{red}{0.502}} &\textbf{\textcolor{blue}{0.446}} &0.403&\textbf{\textcolor{green}{0.493}}&0.298\\
			DP & \textbf{\textcolor{red}{0.712}} &\textbf{\textcolor{blue}{0.612}} &0.532&\textbf{\textcolor{green}{0.678}}&0.390\\
			\bottomrule[1.5pt]			
		\end{tabular}\label{tab:deep_ours}
	\vspace{-0.1cm}
	\end{threeparttable}
\end{table}

\section{Conclusion}
This work puts forward a novel real-time tracker with anti-dark function, \textit{i.e.}, ADTrack. ADTrack first implements image enhancement and target-aware mask generation based on an image enhancer. With the mask, ADTrack innovatively proposes dual filters, \textit{i.e.}, the target-focused filter and the context filter, regression model. Thus, the dual filters restrict each other in training and compensate each other in detection, achieving robust real-time dark tracking onboard UAV. In addition, the first dark tracking benchmark, UAVDark70, is also constructed in this work for visual tracking community. The proposed anti-dark tracker and dark tracking benchmark will make an outstanding contribution to the research of UAV tracking in dark conditions in the future.

%
%

\section*{Acknowledgment}
This work is supported by the National Natural Science Foundation of China (No. 61806148) and Natural Science Foundation
of Shanghai (No. 20ZR1460100).

\bibliographystyle{IEEEtran}
\normalem
\bibliography{ICRA2021}

\end{document}